\documentclass[10pt,twocolumn,letterpaper]{article}

\usepackage[pagenumbers]{cvpr} 

\usepackage{graphicx}
\usepackage{amsmath}
\usepackage{amssymb}
\usepackage{booktabs}

\usepackage{multirow}
\usepackage{booktabs}
\usepackage{threeparttable}

\usepackage[pagebackref,breaklinks,colorlinks]{hyperref}

\usepackage[capitalize]{cleveref}
\crefname{section}{Sec.}{Secs.}
\Crefname{section}{Section}{Sections}
\Crefname{table}{Table}{Tables}
\crefname{table}{Tab.}{Tabs.}

\def\cvprPaperID{10} 
\def\confName{CVPR}
\def\confYear{2024}

\begin{document}

\title{Confidence-Aware RGB-D Face Recognition via Virtual Depth Synthesis}

\author{
Zijian Chen\textsuperscript{1} \\
\and Mei Wang\textsuperscript{1} \\
\and Weihong Deng\textsuperscript{1} \\
\and Hongzhi Shi\textsuperscript{1} \\
\and Dongchao Wen\textsuperscript{1\thanks{Corresponding author.}}  \\
\and Yingjie Zhang\textsuperscript{1} \\
\and Xingchen Cui \textsuperscript{1} \\
\and Jian Zhao\textsuperscript{1} \\
\and {\textsuperscript{1}Inspur (Beijing) Electronic Information Industry Co., Ltd} \\
 {\tt\small\{zijian\_chen, wangmei\_1, dengweihong2020, wendongchao\}@foxmail.com} \\
 {\tt\small\{shihzh, zhangyj-s, cuixingchen, zhao\_jian\}@ieisystem.com}
}
\maketitle

\begin{abstract}
  2D face recognition encounters challenges in unconstrained environments due to varying illumination, occlusion, and pose. Recent studies focus on RGB-D face recognition to improve robustness by incorporating depth information. However, collecting sufficient paired RGB-D training data is expensive and time-consuming, hindering wide deployment. In this work, we first construct a diverse depth dataset generated by 3D Morphable Models for depth model pre-training. Then, we propose a domain-independent pre-training framework that utilizes readily available pre-trained RGB and depth models to separately perform face recognition without needing additional paired data for retraining. To seamlessly integrate the two distinct networks and harness the complementary benefits of RGB and depth information for improved accuracy, we propose an innovative Adaptive Confidence Weighting (ACW). This mechanism is designed to learn confidence estimates for each modality to achieve modality fusion at the score level. Our method is simple and lightweight, only requiring ACW training beyond the backbone models. Experiments on multiple public RGB-D face recognition benchmarks demonstrate state-of-the-art performance surpassing previous methods based on depth estimation and feature fusion, validating the efficacy of our approach.
\end{abstract}

\section{Introduction}
\label{sec:intro}

2D face recognition has shown impressive results with the development of deep learning in recent years. But it still faces challenges such as sensitivity to factors like pose, occlusion, and makeup in open-world scenarios \cite{sepas2020face}. With the increasing convenience of obtaining depth data through consumer depth cameras, more recent studies have focused on RGB-D face recognition. Incorporating additional depth information strengthens RGB-D face recognition methods for handling challenging scenarios and achieving a more robust performance compared to 2D face recognition based only on RGB images \cite{gilani2018Learning}.

\begin{figure*}
    \centering
    \includegraphics[width=0.90\textwidth]{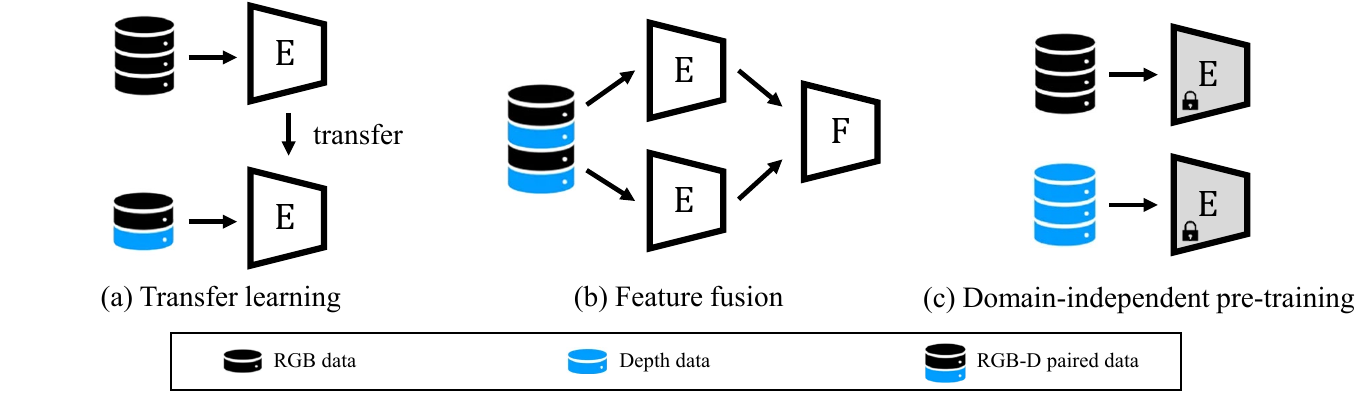}
    \caption{A comparison of our network and other networks.}
    \label{fig:diff}
\end{figure*}

Many methods \cite{zhang2018RGBD,uppal2021two,zhu2023exploiting} have been proposed to enhance the performance of RGB-D face recognition, most of which involve fusing features from the two modalities in the early stage, requiring a large amount of paired RGB-D data for training. However, unlike 2D images that can be easily obtained from the Internet, collecting paired RGB-D data is expensive and time-consuming. To address this issue, some methods \cite{cui2018Improving,chiu2023RGBD} have been proposed to learn a mapping from RGB to depth, which can be used to estimate pseudo depth images. However, according to \cite{akin2022challenges}, estimating depth information from RGB is an ill-posed problem due to the inherent ambiguity between RGB and depth, which can result in the loss of distinctive depth features in the pseudo depth images and limit their recognition performance.

However, the situation changes when we focus solely on depth face data instead of RGB-D paired data. Despite the absence of large-scale depth face recognition datasets, the generation of virtual data appears to be a simpler task. The 3D Morphable Model (3DMM), a classic approach for generating 3D face models, has seen significant advancements since its inception in 1999 \cite{blanz2023morphable}. Over the past two decades, 3DMM has evolved through developments in high-quality face data collection using advanced 3D scanning technologies for parametric modeling \cite{li2017learning, paysan2009face, gerig2018morphable}. These models are now capable of producing highly detailed 3D faces. However, as far as we are aware, no existing method employs 3DMM generated data for face recognition purposes. Therefore, we introduce a method that leverages virtual face data produced by 3DMM, encompassing a broad spectrum of identities, expressions, and poses, for training in deep face recognition. This approach is both straightforward and efficient, significantly reducing the dependency on real data.

Conversely, it is well-known that 2D face recognition is a mature field with a large number of open-source pre-trained models. Consequently, we present a novel yet effective domain-independent pre-training framework. This framework employs readily available pre-trained RGB and depth models in two distinct branches. Diverging from methods that necessitate the fine-tuning of pre-trained models with RGB-D paired data, our approach features fully independent RGB and depth branches. These branches operate without the requirement for joint training on additional paired data, facilitating direct face recognition in RGB and depth images, respectively.

The inherent differences between RGB and depth modalities result in varying discriminative power for face recognition depending on image quality. While RGB excels at texture and color cues, depth sensing provides geometric shape cues that are invariant to lighting and color. To fully exploit the complementary strengths of both modalities, we propose a novel \textbf{A}daptive \textbf{C}onfidence \textbf{W}eight (ACW) to dynamically modulate the contributions of the RGB and depth branches during inference. ACW performs a weighted fusion of similarities between probe and gallery samples from each modality by estimating the confidence of two modalities using a lightweight network. It improves robustness against modal-specific noise and enhances performance in challenging scenarios. Moreover, as a post-processing method during the inference phase, this method can well cope with the absence of modality.

Our contributions are summarized as follows:
\begin{itemize}
    \item We construct a diverse depth dataset generated by 3D Morphable Models for depth model pre-training and propose a simple yet effective domain-independent pre-training framework that enhances the performance of RGB-D face recognition without requiring additional paired training data.
    \item We introduce an innovative adaptive confidence weighting mechanism that adaptively fuses modalities' similarities, improving robustness against modality-specific noise.
    \item We achieve state-of-the-art performance on several public RGB-D face recognition datasets. In particular, on the Lock3DFace dataset, our method obtains an average recognition rate of 97.41\%, outperforming the previous best result by 2.67\%.
\end{itemize}
\section{Related Work}

\begin{figure*}[htbp]
    \centering
    \includegraphics[width=0.9\textwidth]{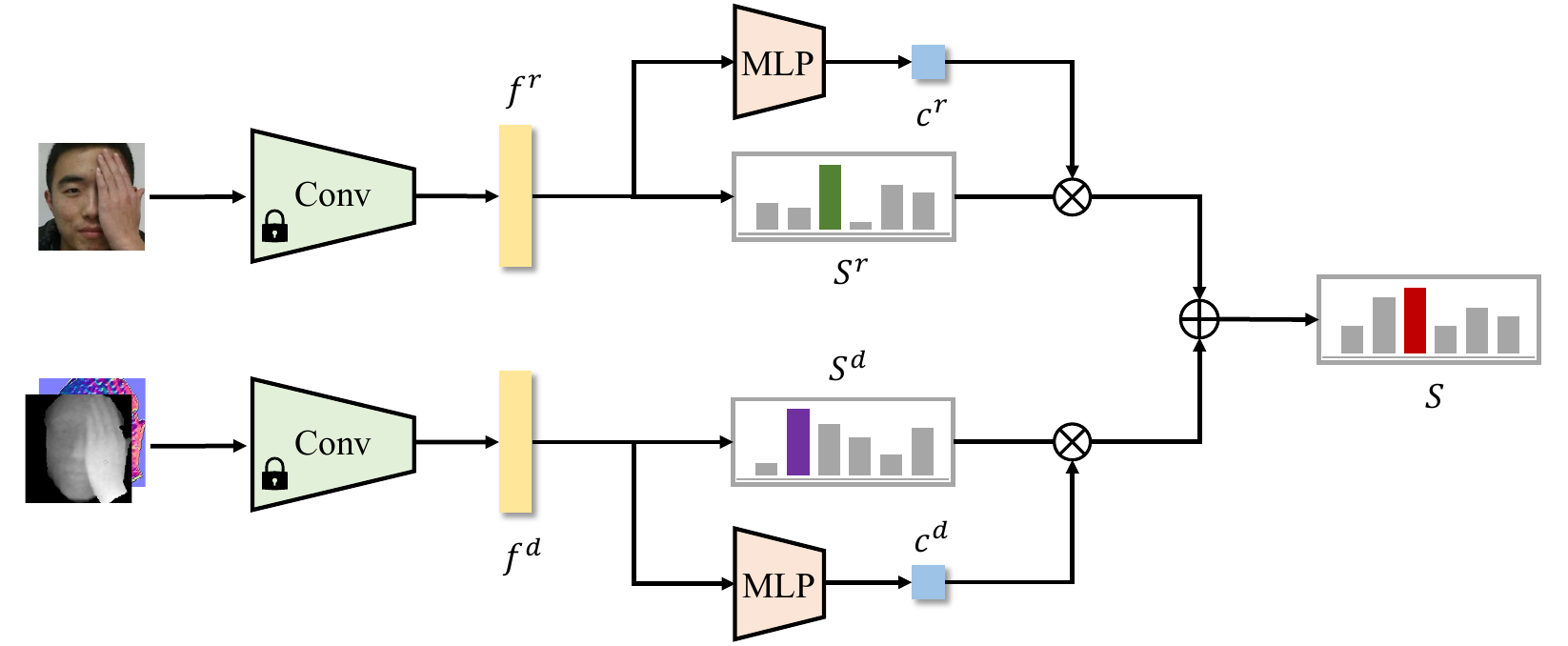}
    \caption{Overview of our proposed method. During training of ACW, two branches freeze the main backbones and are trained completely independently. During inference, RGB and depth are extracted to the features $f_r$ and $f_d$ respectively and fed into MLP to obtain confidence levels $c_r$ and $c_d$. The cosine similarity of the two modalities is weighted and fused using the confidence level, and the final score $S$ is obtained.}
    \label{fig:overview}
\end{figure*}

The main challenges for RGB-D face recognition are obtaining paired RGB-D data and modal fusion strategy. Many studies have proposed methods to address these issues in recent years.

\textbf{Paired RGB-D data.} Prior methods usually generate depth images from RGB images to mitigate the problem of the lack of real paired data. \cite{cui2018Improving} proposed a network architecture of cascaded fully convolutional networks and convolutional neural networks for discriminative depth estimation from 2D face images. The network aims to preserve subject discriminative information in the estimated depth while predicting depth information from an RGB image. \cite{uppal2021TeacherStudent} proposed the Teacher-Student Generative Adversarial Network (TS-GAN) to generate depth images from RGB images across datasets. \cite{chiu2021HighAccuracy} introduced a segmentation-aware depth estimation network that utilizes semantic segmentation information to estimate depth from RGB face images and localize facial regions more accurately. These studies constructed complex and time-consuming models to generate paired RGB-D images but still limit the usage of large amounts of open-source RGB face data. However, our experiments indicate that it is not necessary to use additional paired RGB-D data and that significant improvements can be obtained by open-source RGB face data and a large amount of virtual depth face data without specific pairing.

\textbf{Fusion strategies.} Most of the previous methods use feature-level fusion strategy, while a few utilize signal-level fusion or score-level fusion. \cite{xiong2019improving} only concatenated RGB and depth channels, without considering the differences between the two modalities. \cite{zhang2018RGBD} proposed a score-level fusion method based on the observed matching accuracies by each modality alone. Nevertheless, the fusion weights remained fixed and did not adapt to the varying characteristics of different samples. \cite{chiu2021HighAccuracy} proposed a mask-guided two-stream multi-head network structure, which utilized semantic alignment loss to improve the semantic consistency of two modalities. \cite{uppal2021Depth} proposed an attention module that focused on specific regions of visual features in the RGB image by guiding the network with depth features. \cite{uppal2021two} proposed a two-level attention-based fusion strategy that utilized LSTM recurrent learning to capture the relationship between feature maps and convolution to focus on the spatial features of those maps. \cite{zhu2023exploiting} proposed the masked modeling scheme and iterative inter-modal feature interaction module, aiming to exploit the implicit relations between two modalities. Although these feature fusion methods can achieve good results in modality fusion, they are mostly unable to handle modality missing due to network structure reasons. Furthermore, they heavily depend on joint training between modalities and cannot fully utilize pre-trained 2D face recognition models.

\textbf{Pre-trained models.} The use of pre-trained models has become a widespread practice in the field of 2D face recognition. These models are trained on large-scale datasets. Some studies have explored the use of pre-trained 2D face recognition models for RGB-D face recognition as well. For instance, \cite{xiong2019improving} proposed a method to incorporate depth information by adding a depth channel to the input layer of a pre-trained 2D face recognition network, allowing knowledge transfer from RGB to RGB-D. Other works, such as \cite{zhang2018RGBD,uppal2021Depth,uppal2021two,zhu2023exploiting}, first pre-trained 2D face recognition models as the feature extractor on large 2D face datasets and then fine-tuned their entire models on small RGB-D datasets. Both approaches require fine-tuning on RGB-D data, but this is difficult due to the limited availability of such data, which may lead to catastrophic forgetting and a decline in model performance.

\section{Method}

In this section, we propose our method for RGB-D face recognition. We first introduce our domain-independent pre-training framework in \cref{sec:intro}, which includes the construction of a virtual depth dataset. We then present our Adaptive Confidence Weighting (ACW) mechanism in \cref{sec:acw} for modality fusion at the score level. The overall framework of our proposed method is illustrated in \Cref{fig:overview}.

\subsection{Domain-Independent Pre-training}
\label{sec:pretrain}
RGB-D face recognition methods can be divided into two categories: transfer learning and feature fusion. As shown in \Cref{fig:diff} (a), the method of transfer learning retrains a pre-trained RGB model using a small amount of RGB-D data. Its drawback is requiring paired RGB-D data and not being able to fully adapt to different modalities due to sharing a backbone between the two modalities. As shown in Figure \Cref{fig:diff} (b), the method of feature fusion utilizes two single-modal models to extract features and then fed them into a feature fusion module. This method requires an abundance of paired RGB-D data to train the fusion module, regardless of whether the feature extractors are pre-trained. Meanwhile, the feature fusion module will increase the overhead of the network. In contrast, our proposed domain-independent pre-training has the following advantages: (1) it allows the use of different network architectures for the two modalities; (2) it does not require paired RGB-D data, allowing the use of single-modal virtual depth data for pre-training; (3) the backbones don't need to be retrained or fine-tuned, thus avoiding additional computational overhead.

As shown in \Cref{fig:overview}, we use two separate networks to perform recognition for RGB and depth images. The two networks which pre-training independently can have different structures without the need for paired RGB-D data. The modality fusion is performed at the score level during inference, thus there is no need for retraining.

\textbf{Pre-trained model.} We choose two lightweight models for domain-independent pre-training. For the RGB domain, we directly use the MobileFaceNet \cite{chen2018mobilefacenets} model pre-trained on the CASIA Webface dataset \cite{yi2014learning}. For the depth domain, there is no open-source model pre-trained on large-scale depth datasets available. Therefore, we propose a method to construct a large-scale virtual depth dataset by 3DMM.

\begin{figure}
    \centering
    \includegraphics[width=0.95\linewidth]{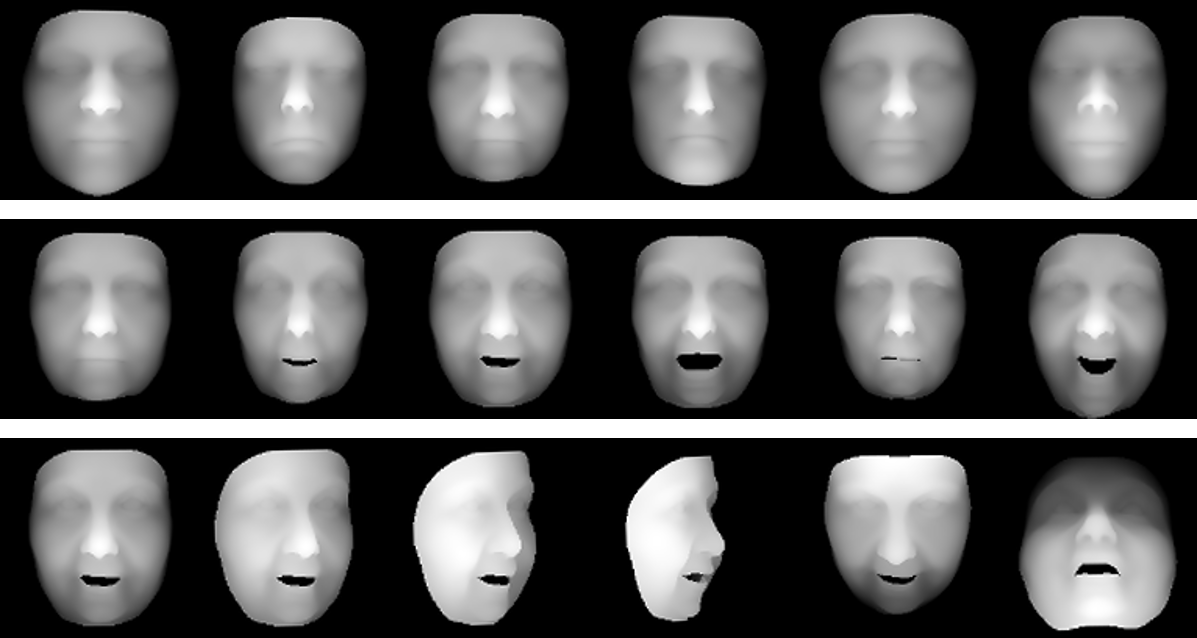}
    \caption{Generated depth images showing variations in identity, expression, and pose. Each row represents samples with different identities, expressions, and poses, respectively.}
    \label{fig:bfm}
\end{figure}

We use the Led3D \cite{mu2019led3d} model with inputs of depth and normal images as the backbone. Since paired data is not required, we don't need to generate depth data from the CASIA Webface dataset. Instead, we can directly use the \textit{BFM 2019}\cite{gerig2018morphable} to generate virtual face data. It models the 3D face as a linear combination of a mean face, identity, and expression variations. The 3D face shape $\mathbf S$ is represented as:

\begin{equation}
    \mathbf S = \overline{\mathbf S} + \mathbf A_{id} \mathbf \alpha_{id} + \mathbf A_{exp} \mathbf \alpha_{exp}
\end{equation}

where $\overline{\mathbf S}$ is the mean face, $\mathbf A_{id}$ and $\mathbf A_{exp}$ are the identity and expression bases, and $\mathbf \alpha_{id}$ and $\mathbf \alpha_{exp}$ are the identity and expression coefficients. By randomly sampling the coefficients, we can generate a large number of virtual high-quality faces with different identities and expressions. To introduce pose variations and large occlusions in each 3D scan, we simulate these by deploying 12 synthetic cameras on a hemisphere in front of the 3D face. Some images generated by our method are shown in \Cref{fig:bfm}. Despite its simplicity, this method of data generation significantly enhances the efficacy of face recognition technologies. 
We generated 10,000 identities, each with 1 neutral expression and 40 random expressions, and each face corresponds to 12 different poses, resulting in a total of 4,920,000 images. As shown in \Cref{tab:dataset}, the number of identities is nearly 20 times that of Lock3DFace, the largest RGB-D face recognition dataset, and is comparable to the RGB face dataset CASIA Webface.

We first pre-train the depth model on high-quality virtual data as $model_{HQ}$. It can be used to infer high-quality depth images, e.g., on the Bosphorus dataset. However, in real-world scenarios, low-quality depth images captured by consumer depth cameras have significant discrepancies compared to virtual data. To address this issue, we fine-tune $model_{HQ}$ using the low-quality dataset, Lock3DFace, which enables it to adapt to the distribution of low-quality 3D faces. The resulting fine-tuned model, $model_{LQ}$, is more effective in handling the complexities of low-quality depth images.

\textbf{Modality fusion.} During inference, we first obtain the features from each modality and calculate the cosine similarities $s$ between the probe and gallery images for each modality. We then perform modality fusion at the score layer based on the similarities. The weights for each modality's similarity score $c\in[0, 1]$ are controlled by their corresponding confidences. Finally, the weighted similarities are summed to obtain the final score:

\begin{equation} \label{eq:score}
    s_i = \sum_{j=1}^{M}{c^j \cdot s^j_i}
\end{equation}
where $s^j_i$ represents the similarity between the sample and the $i^{th}$ identity of the $j^{th}$ modality, and $c^j$ represents the confidence of the $j^{th}$ modality. The highest matching score is chosen as the final matching result for the probe.


\begin{table}
    \centering
    \begin{threeparttable}[b]
        \begin{tabular}{lcc}
            \toprule
            Dataset & \# Identities & \# Images \\
            \midrule
        Lock3DFace & 509 & 334,589\tnote{1} \\
        CASIA Webface & 10,575 & 494,414 \\
        \textbf{Our virtual dataset} & \textbf{10,000} & \textbf{4,920,000} \\
        \bottomrule
    \end{tabular}
    \small
    \begin{tablenotes}
        \item[1] Obtained from 5,671 nearly static video sequences.
    \end{tablenotes}
    \end{threeparttable}
    \caption{Comparison of the scale of our virtual dataset, Lock3DFace, and CASIA Webface.}
    \label{tab:dataset}
\end{table}

\subsection{Adaptive Confidence Weighting}
\label{sec:acw}
To achieve superior fusion, the value of confidence is crucial. While the confidence of each modality can be manually set as a constant value, it may vary for different samples, particularly those with large poses or occlusions. Therefore, we propose a lightweight Adaptive Confidence Weighting (ACW) mechanism, which takes the feature extracted by the pre-trained networks of each modality as input, to adaptively learn the confidence of each modality for each sample. Inspired by \cite{devries2018learning}, ACW consists of a sigmoid-activated MLP that outputs a confidence value $c$. The value $c$ represents the level of confidence in the network's ability to correctly predict the identity, where values closer to 1 indicate higher confidence, and vice versa.

To effectively learn the same meaningful confidence across multiple modalities, we introduce the updated logits $z'$ obtained by interpolating between the single modality network's predicted logits $z$ and the ground-truth label $y$, controlled by the confidence of the network. This allows ACW to use the updated logits for single modality classification. The interpolation is performed as follows:
\begin{equation} \label{eq:pythagoras}
    z'^j_i = c^j \cdot z^j_i + (1 - c^j) y_i
\end{equation}
where $z'^j_i$ represents the updated logit for the $i^{th}$ identity for the $j^{th}$ modality. We employ the cross-entropy loss as the task loss for classification:
\begin{equation} \label{eq:celoss}
    \mathcal{L}^j_t = - \sum_{i=1}^{N} {\log(p'^j_i)y_i}
\end{equation}
where $p'^j_i$ is obtained by applying softmax to $z'^j_i$. 

A confidence of 0 indicates that the network's output is entirely unreliable and relies solely on the ground truth. Conversely, a confidence of 1 indicates that the network's output is completely trustworthy and independent of any ground-truth cues. It is worth noting that unlike \cite{devries2018learning} which interpolates probabilities, we use logits since face recognition is an open-set problem, and inference does not involve a softmax operation to obtain probabilities, but rather a cosine similarity between probe and gallery images. The similarity and logits during training are consistent.

In addition to the task loss, a confidence loss is also required to encourage the network to output higher confidence when it is confident about its prediction and to prevent the confidence from approaching zero, excessively relying on ground truth labels. We use the negative log-likelihood of the confidence as the confidence loss:
\begin{equation} \label{eq:confloss}
    \mathcal{L}^j_c = - \log(c^j)
\end{equation}
The total loss function of ACW is the weighted sum of the task loss and the confidence loss for each modality:
\begin{equation} \label{eq:totalloss}
    \mathcal{L} = \sum_{j=1}^{M} {\mathcal{L}^j_t + \lambda \cdot \mathcal{L}^j_c}
\end{equation}
Training of ACW requires real RGB-D datasets. However, since the module is very lightweight, it can learn the confidence distribution with only a small amount of data. Meanwhile, ACW is a simple yet effective method that can be easily integrated into any multimodal fusion network, not just limited to RGB-D face recognition.

\section{Experiments}

\begin{table*}[htbp]
    \centering
    \begin{tabular}{c|c|c|cccccc}
    \toprule
    Method & Model type & Input & NU & FE & PS & OC & TM & Total \\
    \midrule
  \cite{mu2019led3d} & \multirow{4}[0]{*}{Depth} & Depth + Normal & 99.62 & 98.17 & 70.38 & 78.10 & 65.28 & 84.22 \\
  \cite{lin2021high} && Normal & 99.95 & 97.31 & 73.61 & 80.97 & 61.67 & 86.55 \\
  \cite{zhao2022LMFNet} && Depth + Normal  & 99.95 & 99.21 & 79.05 & 84.40 & 76.31 & 88.01 \\
  \textbf{Ours} && Depth + Normal & \textbf{99.95} & \textbf{99.45} & \textbf{84.92} & \textbf{91.59} & \textbf{77.54} & \textbf{90.48} \\
    \midrule
  \cite{cui2018Improving} & \multirow{6}[0]{*}{RGB-D} & RGB + Depth & - & 97.30 & 54.6 & 69.60 & 66.10 & 71.90 \\
  \cite{uppal2021Depth} && RGB + Depth & - & 99.40 & 70.60 & 85.80 & 81.10 & 87.30 \\
  \cite{zhu2023exploiting} && RGB + Depth & - & 99.50 & 75.60 & 90.30 & 85.26 & 90.60 \\
  \cite{chiu2023RGBD} && RGB + Depth + Mask & - & 99.69 & 93.89 & 93.13 & \textbf{91.86} & 94.74 \\
  \textbf{Ours} && RGB + Depth + Normal & - & \textbf{100.00} & \textbf{98.86} & \textbf{99.40} & 90.03 & \textbf{97.41} \\
    \bottomrule
    \end{tabular}
    \caption{Rank-1 recognition rate (\%) comparison on the Lock3DFace dataset}
    \label{tab:comparison}
\end{table*}

\subsection{Datasets and Protocols}
We evaluate our proposed method on three public RGB-D face recognition datasets: Lock3DFace \cite{zhang2016lock3dface}, IIIT-D \cite{goswami2013RGBD}, and Bosphorus \cite{savran2008bosphorus}.

\textbf{Lock3DFace.} Lock3DFace is one of the largest low-quality RGB-D face dataset public available to date, containing 5,671 video sequences of 509 individuals. Each sequence consists of 59 frames of images captured by the Kinect v2 sensor in an indoor environment with natural lighting. Additionally, this dataset encompasses the most comprehensive facial changes in real life, including neutral, expression, pose, occlusion, and time changes. 
We follow the setting in \cite{cui2018Improving} that divides the training and testing sets based on subjects. We randomly select 340 subjects as the training set, and the remaining 169 subjects as the testing set. The testing set is further divided into 5 subsets, namely ``NU'', ``FE'', ``PS'', ``OC'', and ``TM'', as described in \cite{zhang2016lock3dface}.

\textbf{IIIT-D.} The IIIT-D RGB-D dataset is a low-quality RGB-D face dataset that contains 106 subjects with 4,603 images captured by the Kinect v1 sensor in two acquisition sessions. The dataset contains variations in pose, expression, and eyeglasses. All images are pre-cropped around the face. We strictly follow the predefined protocol in \cite{goswami2013RGBD} with a five-fold cross-validation strategy, where 4 samples are used as the gallery and the remaining as the probe for each subject in each fold. 

\textbf{Bosphorus.} The Bosphorus dataset is a high-quality RGB-D face dataset that includes 4,561 images of 105 subjects with various poses, expressions, and occlusions. Following the protocol in \cite{chiu2023RGBD}, we use the first neutral image of each subject to form the gallery set with 105 images and the remaining images for identification.

\subsection{Implementation Details}
To pre-train the depth model, we utilize the Adam optimizer with learning rate $\gamma = 0.0003$. We train the model using 384 samples per batch for 2 epochs on virtual depth images and 20 epochs on real depth images. The input image size for the depth model is $128\times128$, following the setting in \cite{mu2019led3d}. The normal images are generated from their corresponding depth images. The output feature length for both the RGB and depth models is 512. Additionally, both the RGB and depth models use the ArcFace \cite{Deng2018ArcFace} loss function.

When training ACW, we freeze the feature extractors and train ACW on Lock3DFace with a batch size of 384 samples for 20 epochs. We use SGD as the optimizer with learning rate $\gamma = 0.006$. To ensure that the logits during training are close to the cosine similarity during inference, we initialize the weights of the classifier with the features of a neutral image of each class in the training set. The setting of $\lambda$ in \cref{eq:totalloss} is consistent with \cite{devries2018learning}. All experiments are conducted on a single NVIDIA RTX 3090 GPU.

\subsection{Performance and comparison} To validate our proposed method, we compare it with other methods on the Lock3DFace dataset. As shown in \Cref{tab:comparison}, the results show that our method achieves state-of-the-art performance for both tasks.

With depth as input, our model architecture is similar to \cite{mu2019led3d} except for the output dimension. However, our average recognition rate is significantly improved, which validates the effectiveness of using generated virtual depth images for pre-training. Although there is a quality difference between the virtual data and real data, the virtual data provides prior knowledge that enables the model to better understand faces. Compared to the second-best method, our average recognition rate is increased by 2.47\%, and we outperform all other methods in every subset, especially in pose and occlusion subsets, where we achieved a 5.87\% and 7.19\% improvement, respectively. This is mainly due to our virtual data containing multiple viewpoints, including some large poses and self-occlusions. It shows that the simple pipeline of virtual data generation can significantly improve the performance of the depth model.

With RGB and depth as input, our method also excels. Our method further improves the recognition rate by 0.62\% to 97.41\%. It demonstrates the effectiveness of domain-independent pre-train with virtual depth data and the ACW mechanism. Compared to previous state-of-the-art methods \cite{chiu2023RGBD} which generate depth images from the VGGFace2 dataset, one magnitude larger than our CASIA WebFace dataset, and use semantic segmentation masks for RGB-D joint training, we achieve the best performance on almost all subsets and improve the total recognition rate by 2.67\%. This indicates that inaccurate depth estimation may weaken performance. While our method achieves better performance by utilizing pre-trained models with a straightforward architecture, avoiding the need for complex feature fusion that requires paired RGB-D data for training.

As shown in \Cref{tab:others}, we further evaluate the generalization performance of our RGB-D face recognition method on other testing datasets without fine-tuning. Specifically, we employ the same depth model $model_{LQ}$ as in Lock3DFace to test our method on the IIIT-D dataset, achieving a recognition rate of 99.7\%, which is consistent with the state-of-the-art performance in \cite{uppal2021Depth}. For the high-quality Bosphorus dataset, we use $model_{HQ}$ as the depth model and achieve a recognition rate of 97.9\%, outperforming the previous state-of-the-art method \cite{chiu2023RGBD} by a slight margin of 0.3\%. This demonstrates that our method is effective even on cross-quality datasets and can achieve comparable or better performance by simply replacing the depth model.

\begin{table}
    \centering
    \begin{tabular}{ccc}
    \toprule
        Dataset & Method & Accuracy \\
    \midrule
        \multirow{5}[0]{*}{IIIT-D} &
        \cite{chowdhury2016rgb} & 98.7 \\
        & \cite{cui2018Improving} & 96.5 \\
        & \cite{uppal2021two} & 99.4 \\
        & \cite{uppal2021Depth} & \textbf{99.7} \\
        & \textbf{Ours} & \textbf{99.7} \\
    \midrule
        \multirow{5}[0]{*}{Bosphorus} &
        \cite{li2015towards} & 96.6 \\
        & \cite{mian2007Efficient} & 96.4 \\
        & \cite{gilani2018Learning} & 96.2 \\
        & \cite{chiu2023RGBD} & 97.6 \\
        & \textbf{Ours} & \textbf{97.9} \\
    \bottomrule
    \end{tabular}
    \caption{Rank-1 recognition rate (\%) comparison on the IIIT-D and Bosphorus datasets. The data for the Bosphorus dataset is from \cite{chiu2023RGBD}.}
    \label{tab:others}
\end{table}

\begin{table*}
    \centering
    \begin{tabular}{cccc|ccccc}
    \toprule
    Model & RGB pre-trained & Depth pre-trained & ACW & FE & PS & OC & TM & Total \\
    \midrule
    $M_1$ & & & & 99.25 & 74.85 & 91.99 & 60.91 & 84.88 \\
    $M_2$ && & $\surd$ & 99.32 & 81.94 & 91.74 & 73.04 & 88.91 \\
    $M_3$ && $\surd$ & & 99.49 & 82.04 & 93.94 & 68.09 & 88.23 \\
    $M_4$ &$\surd$ & & & 100.00 & 98.07 & 98.95 & 89.60 & 97.08 \\
    $M_5$ &$\surd$ & $\surd$ & & 99.96 & 98.26 & 99.04 & 89.95 & 97.20 \\
    $M_6$ &$\surd$ & $\surd$ & $\surd$ & \textbf{100.00} & \textbf{98.86} & \textbf{99.40} & \textbf{90.03} & \textbf{97.41} \\
    \bottomrule
    \end{tabular}
    
    \caption{Comparison in terms of rank-1 recognition rate (\%) on Lock3DFace with ablations in our proposed method. $M_1$ is the baseline model without pre-training and ACW. $M_2$ is the model with ACW. $M_3$ and $M_4$ are the models pre-trained on depth and RGB images, respectively. $M_5$ is the model pre-trained on both depth and RGB images. $M_6$ is our proposed model.}
    \label{tab:ablation}
\end{table*}

\subsection{Ablation study} We conduct ablation experiments to validate the effectiveness of two key components of our proposed method: (1) pre-training on large RGB and depth datasets, and (2) the adaptive confidence weighting. We train six models with different configurations, as summarized in \Cref{tab:ablation}. $M_1$ is a baseline model that was trained from scratch on Lock3DFace without any pre-training. It directly combines two similarity scores for modality fusion without utilizing ACW. $M_2$ adds ACW to the baseline. $M_3$ and $M_4$ are pre-trained on depth and RGB datasets, respectively. $M_5$ incorporates both RGB and depth pre-training. Finally, $M_6$ is our full proposed model.

First, we evaluate the effect of ACW by comparing the recognition rates of the baseline model and the model with ACW ($M_1$ and $M_2$, respectively). The results show that $M_2$ outperforms $M_1$ by 4.02\%, which indicates that incorporating effective similarity fusion can significantly improve performance. Then we compare the performance of the model without ACW and the proposed method with ACW ($M_5$ and $M_6$, respectively). The results demonstrate a 0.21\% improvement over a baseline of 97.20\% achieved through pre-training.  Further experimental validation of ACW will be presented in the following evaluation of challenging scenarios.

In addition, our network architecture allows pre-training on large single-modality RGB and depth datasets. To validate the importance of large-scale pre-training, we conduct ablation experiments of pre-training by replacing the large-scale dataset with the Lock3DFace dataset on RGB and depth, respectively. $M_3$ is pre-trained only on depth, while $M_4$ is pre-trained only on RGB. We observe that $M_4$ outperforms the baseline model by 12.20\%, which suggests that a pre-trained RGB model can significantly improve the accuracy. Meanwhile, pre-training on depth results in a good performance improvement as well, although not as significant as pre-trained on RGB, improving the rank-1 recognition rate by 3.35\%. Furthermore, comparing $M_2$ which uses only ACW with the proposed method $M_6$, we see an improvement of 8.5\% in rank-1 recognition rate. This shows that pre-training on both modalities is crucial for achieving high accuracy in the final model.

\begin{figure}
    \centering
    \includegraphics[width=0.7\linewidth]{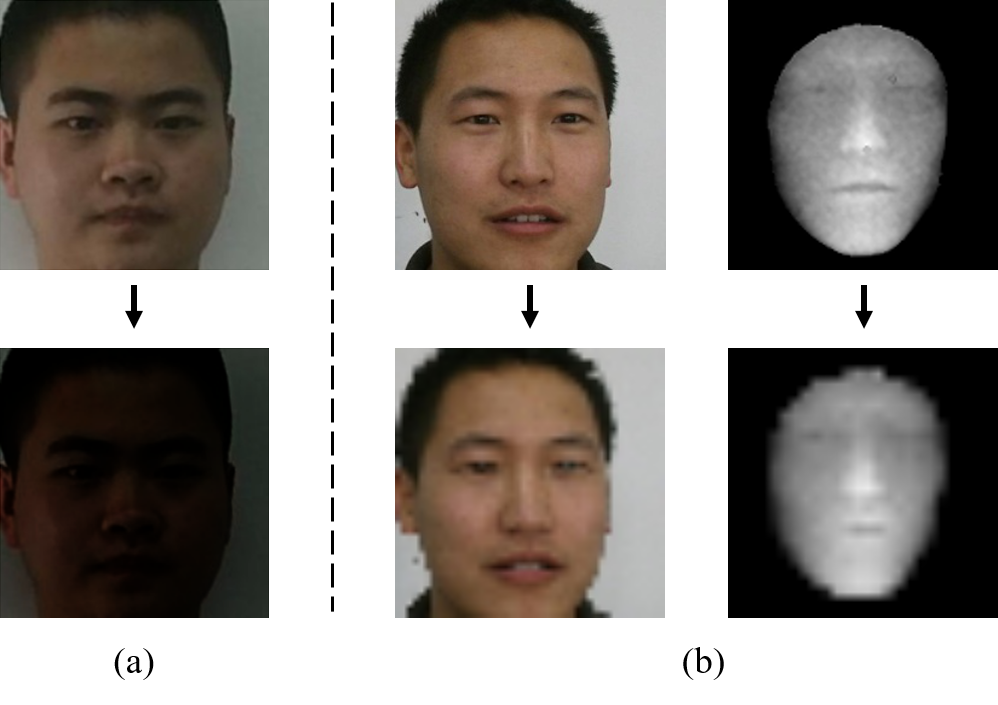}
    \caption{Simulating challenging scenarios. (a) Gamma correction algorithm simulates low-light scenarios. (b) Scaling down the images simulates far-distance scenarios.}
    \label{fig:challenging}
\end{figure}

\subsection{Evaluation on challenging scenarios} RGB-D face recognition is more suitable for a variety of challenging scenarios compared to 2D face recognition. To demonstrate the robustness of ACW in tackling challenging scenarios, we design two experiments: one on low-light conditions and the other on distant faces. The baseline model proportionally fuses similarities for RGB and depth.

In the first experiment, we simulate low-light scenarios using the gamma correction algorithm as shown in \Cref{fig:challenging} (a). As shown in \Cref{tab:light}, when $\gamma$ is set to 0.3, 0.4, and 0.6, ACW slightly improves the recognition rate by 0.20\%, 0.35\%, and 0.23\%, respectively, compared to the baseline model. There is literature evidence that neural networks have certain robustness to changes in lighting, which may be the reason for the relatively weak improvement. Nevertheless, ACW consistently outperforms the baseline model in multiple experiments, confirming its ability to enhance robustness to lighting changes.

In the second experiment, we use scaling algorithms with reduced RGB and depth image resolution to simulate faces that are farther away as shown in \Cref{fig:challenging} (b). As shown in \Cref{tab:distance}, ACW also outperforms the baseline model under different scaling factors $f$, with all subsets showing improvement. Specifically, the improvement is more significant when $f$ is small, with an average recognition rate increase of 3.99\% when $f=0.2$. This indicates that ACW can significantly improve the robustness of the model to face distance.

These results indicate that ACW is effective in adapting to different scenarios with varying image qualities in two modalities.

\begin{table}
    \centering
    \small
    \begin{tabular}{ccccccc}
    \toprule
        Method & $\gamma$ & FE & PS & OC & TM & Total \\
    \midrule
        baseline & 0.3 & 99.96 & 95.73 & 98.30 & 84.67 & 95.41 \\
        ACW & 0.3 & 100.00 & 96.73 & 98.80 & 84.36 & 95.61 \\
    \midrule
        baseline & 0.4 & 99.96 & 97.17 & 98.84 & 88.57 & 96.65 \\
        ACW & 0.4 & 100.00 & 98.21 & 99.19 & 88.96 & 97.00 \\
    \midrule
        baseline & 0.6 & 99.96 & 97.92 & 99.05 & 89.72 & 97.09 \\
        ACW & 0.6 & 100.00 & 98.71 & 99.25 & 89.87 & 97.32 \\
    \bottomrule
    \end{tabular}
    \caption{Results in challenging scenario 1, simulated low-light conditions with different gamma coefficients.}
    \label{tab:light}
\end{table}

\begin{table}
    \centering
    \small
    \begin{tabular}{ccccccc}
    \toprule
    Method & $f$ & FE & PS & OC & TM & Total \\
    \midrule
        baseline & 0.2 & 94.94 & 78.91 & 75.62 & 64.02 & 82.06 \\
        ACW & 0.2 & 98.38 & 82.68 & 77.32 & 72.65 & 86.05 \\
    \midrule
        baseline & 0.4 & 99.96 & 96.63 & 98.35 & 88.73 & 96.50 \\
        ACW & 0.4 & 100.00 & 98.26 & 99.10 & 89.87 & 97.21 \\
    \midrule
        baseline & 0.6 & 99.96 & 97.57 & 99.05 & 89.80 & 97.05 \\
        ACW & 0.6 & 100.00 & 98.66 & 99.40 & 90.07 & 97.38 \\
    \bottomrule
    \end{tabular}
    \caption{Results in challenging scenario 2, scaling images with different scaling coefficients to simulate faces at a distance.}
    \label{tab:distance}
\end{table}

\begin{figure}
    
    \centering
    \includegraphics[width=\linewidth]{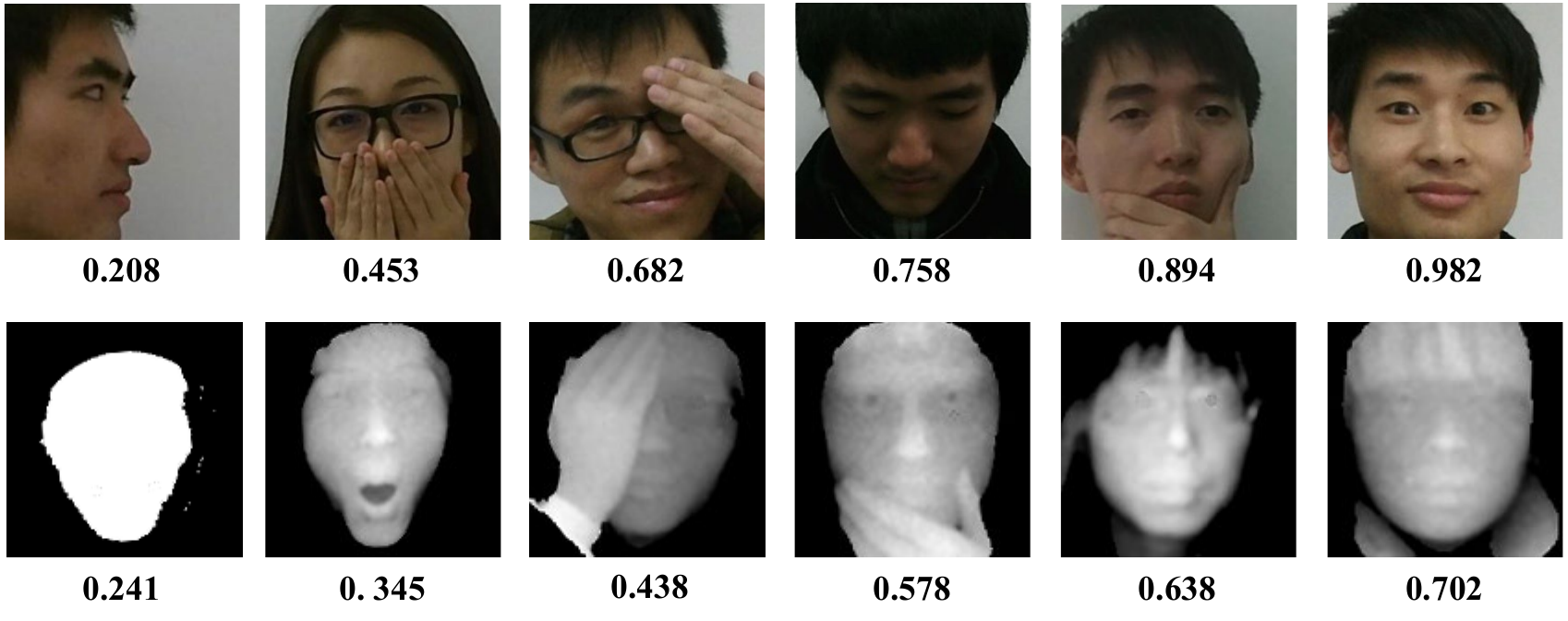}
    \caption{The confidence of the two modalities obtained by our method on the Lock3DFace dataset. It demonstrates a strong correlation between confidence and image quality.}
    \label{fig:confidence}
\end{figure}

\subsubsection{Visualization of confidences} To further demonstrate the effectiveness of our ACW, we visualize the confidences for some samples in Lock3DFace. As shown in \Cref{fig:confidence}, the confidences of the two modalities are consistent with the image quality. When the image quality is poor, such as occlusion or large pose, the confidence will decrease. This shows that the network has learned the confidence of the two modalities and can distinguish the image quality. It is worth noting that the confidence of depth is usually lower than that of RGB. This is consistent with our expectations because the quality of depth is usually worse than that of RGB. However, the difference in confidence between RGB and depth varies for different samples. Therefore, our ACW can adaptively adjust the similarity of the two modalities based on the difference in confidence.

\section{Conclusion}
In this work, we have introduced a method for virtual depth image generation, a simple domain-independent pre-training framework and the adaptive confidence weighting for RGB-D face recognition. Our method achieves state-of-the-art performance on several RGB-D face recognition datasets. The large-scale virtual depth dataset generated by our method shows amazing improvements in depth recognition and RGB-D recognition.
The pre-training framework advances the field by leveraging large-scale pre-trained models without requiring paired data. This is not only more practical but also avoids fine-tuning that could degrade pre-trained models. The lightweight ACW further complements our method by learning modalities' confidences and intelligently fusing similarities. ACW improves robustness, especially in challenging scenarios with noise and quality variations. Overall, this work provides an effective RGB-D face recognition method suitable for real-world applications.

{\small
\bibliographystyle{ieee_fullname}
\bibliography{egbib}

\begin{thebibliography}{10}\itemsep=-1pt

\bibitem{akin2022challenges}
Zeynep Ak{\i}n and Ahmet Sayar.
\newblock Challenges in determining the depth in 2-d images.
\newblock In {\em 2022 International Conference on INnovations in Intelligent SysTems and Applications (INISTA)}, pages 1--6. IEEE, 2022.

\bibitem{blanz2023morphable}
Volker Blanz and Thomas Vetter.
\newblock A morphable model for the synthesis of 3d faces.
\newblock In {\em Seminal Graphics Papers: Pushing the Boundaries, Volume 2}, pages 157--164. 1999.

\bibitem{chen2018mobilefacenets}
Sheng Chen, Yang Liu, Xiang Gao, and Zhen Han.
\newblock Mobilefacenets: Efficient cnns for accurate real-time face verification on mobile devices.
\newblock In {\em Biometric Recognition: 13th Chinese Conference, CCBR 2018, Urumqi, China, August 11-12, 2018, Proceedings 13}, pages 428--438. Springer, 2018.

\bibitem{chiu2021HighAccuracy}
Meng-Tzu Chiu, Hsun-Ying Cheng, Chien-Yi Wang, and Shang-Hong Lai.
\newblock High-accuracy rgb-d face recognition via segmentation-aware face depth estimation and mask-guided attention network.
\newblock In {\em 2021 16th IEEE International Conference on Automatic Face and Gesture Recognition (FG 2021)}, pages 1--8, Dec. 2021.

\bibitem{chiu2023RGBD}
Meng-Tzu Chiu, Hsun-Ying Cheng, Chien-Yi Wang, and Shang-Hong Lai.
\newblock Rgb-d face recognition with identity-style disentanglement and depth augmentation.
\newblock {\em IEEE Transactions on Biometrics, Behavior, and Identity Science}, pages 1--1, 2023.

\bibitem{chowdhury2016rgb}
Anurag Chowdhury, Soumyadeep Ghosh, Richa Singh, and Mayank Vatsa.
\newblock Rgb-d face recognition via learning-based reconstruction.
\newblock In {\em 2016 IEEE 8th International Conference on Biometrics Theory, Applications and Systems (BTAS)}, pages 1--7. IEEE, 2016.

\bibitem{cui2018Improving}
Jiyun Cui, Hao Zhang, Hu Han, Shiguang Shan, and Xilin Chen.
\newblock Improving 2d face recognition via discriminative face depth estimation.
\newblock In {\em 2018 International Conference on Biometrics (ICB)}, pages 140--147, Feb. 2018.

\bibitem{Deng2018ArcFace}
Jiankang Deng, J. Guo, J. Yang, Niannan Xue, Irene Kotsia, and Stefanos Zafeiriou.
\newblock Arcface: Additive angular margin loss for deep face recognition.
\newblock {\em IEEE TPAMI}, 44:5962--5979, 2018.

\bibitem{devries2018learning}
Terrance DeVries and Graham~W Taylor.
\newblock Learning confidence for out-of-distribution detection in neural networks.
\newblock {\em arXiv preprint arXiv:1802.04865}, 2018.

\bibitem{gerig2018morphable}
Thomas Gerig, Andreas Morel-Forster, Clemens Blumer, Bernhard Egger, Marcel Luthi, Sandro Sch{\"o}nborn, and Thomas Vetter.
\newblock Morphable face models-an open framework.
\newblock In {\em 2018 13th IEEE International Conference on Automatic Face \& Gesture Recognition (FG 2018)}, pages 75--82. IEEE, 2018.

\bibitem{gilani2018Learning}
Syed~Zulqarnain Gilani.
\newblock Learning from millions of 3d scans for large-scale 3d face recognition.
\newblock In {\em CVPR}, pages 1896--1905, June 2018.

\bibitem{goswami2013RGBD}
Gaurav Goswami, Samarth Bharadwaj, Mayank Vatsa, and Richa Singh.
\newblock On rgb-d face recognition using kinect.
\newblock In {\em 2013 IEEE Sixth International Conference on Biometrics: Theory, Applications and Systems (BTAS)}, pages 1--6. IEEE, 2013.

\bibitem{li2015towards}
Huibin Li, Di Huang, Jean-Marie Morvan, Yunhong Wang, and Liming Chen.
\newblock Towards 3d face recognition in the real: a registration-free approach using fine-grained matching of 3d keypoint descriptors.
\newblock {\em IJCV}, 113:128--142, 2015.

\bibitem{li2017learning}
Tianye Li, Timo Bolkart, Michael~J Black, Hao Li, and Javier Romero.
\newblock Learning a model of facial shape and expression from 4d scans.
\newblock {\em ACM TOG}, 36(6):194--1, 2017.

\bibitem{lin2021high}
Shisong Lin, Changyuan Jiang, Feng Liu, and Linlin Shen.
\newblock High quality facial data synthesis and fusion for 3d low-quality face recognition.
\newblock In {\em 2021 IEEE International Joint Conference on Biometrics (IJCB)}, pages 1--8. IEEE, 2021.

\bibitem{mian2007Efficient}
Ajmal Mian, Mohammed Bennamoun, and Robyn Owens.
\newblock An efficient multimodal 2d-3d hybrid approach to automatic face recognition.
\newblock {\em IEEE TPAMI}, 29(11):1927--1943, Nov. 2007.

\bibitem{mu2019led3d}
Guodong Mu, Di Huang, Guosheng Hu, Jia Sun, and Yunhong Wang.
\newblock Led3d: A lightweight and efficient deep approach to recognizing low-quality 3d faces.
\newblock In {\em Proceedings of the IEEE/CVF Conference on Computer Vision and Pattern Recognition}, pages 5773--5782, 2019.

\bibitem{paysan2009face}
Pascal Paysan, Marcel L{\"u}thi, Thomas Albrecht, Anita Lerch, Brian Amberg, Francesco Santini, and Thomas Vetter.
\newblock Face reconstruction from skull shapes and physical attributes.
\newblock In {\em Pattern Recognition: 31st DAGM Symposium, Jena, Germany, September 9-11, 2009. Proceedings 31}, pages 232--241. Springer, 2009.

\bibitem{savran2008bosphorus}
Arman Savran, Ne{\c{s}}e Aly{\"u}z, Hamdi Dibeklio{\u{g}}lu, Oya {\c{C}}eliktutan, Berk G{\"o}kberk, B{\"u}lent Sankur, and Lale Akarun.
\newblock Bosphorus database for 3d face analysis.
\newblock In {\em Biometrics and Identity Management: First European Workshop, BIOID 2008, Roskilde, Denmark, May 7-9, 2008. Revised Selected Papers 1}, pages 47--56. Springer, 2008.

\bibitem{sepas2020face}
Alireza Sepas-Moghaddam, Fernando~M Pereira, and Paulo~Lobato Correia.
\newblock Face recognition: a novel multi-level taxonomy based survey.
\newblock {\em IET Biometrics}, 9(2):58--67, 2020.

\bibitem{uppal2021Depth}
Hardik Uppal, Alireza {Sepas-Moghaddam}, Michael Greenspan, and Ali Etemad.
\newblock Depth as attention for face representation learning.
\newblock {\em IEEE Transactions on Information Forensics and Security}, 16:2461--2476, 2021.

\bibitem{uppal2021TeacherStudent}
Hardik Uppal, Alireza {Sepas-Moghaddam}, Michael Greenspan, and Ali Etemad.
\newblock Teacher-student adversarial depth hallucination to improve face recognition.
\newblock In {\em ICCV}, pages 3671--3680, 2021.

\bibitem{uppal2021two}
Hardik Uppal, Alireza Sepas-Moghaddam, Michael Greenspan, and Ali Etemad.
\newblock Two-level attention-based fusion learning for rgb-d face recognition.
\newblock In {\em ICPR}, pages 10120--10127. IEEE, 2021.

\bibitem{xiong2019improving}
Xingwang Xiong, Xu Wen, and Cheng Huang.
\newblock Improving rgb-d face recognition via transfer learning from a pretrained 2d network.
\newblock In {\em International Symposium on Benchmarking, Measuring and Optimization}, pages 141--148. Springer, 2019.

\bibitem{yi2014learning}
Dong Yi, Zhen Lei, Shengcai Liao, and Stan~Z Li.
\newblock Learning face representation from scratch.
\newblock {\em arXiv preprint arXiv:1411.7923}, 2014.

\bibitem{zhang2018RGBD}
Hao Zhang, Hu Han, Jiyun Cui, Shiguang Shan, and Xilin Chen.
\newblock Rgb-d face recognition via deep complementary and common feature learning.
\newblock In {\em 2018 13th IEEE International Conference on Automatic Face \& Gesture Recognition (FG 2018)}, pages 8--15. IEEE, 2018.

\bibitem{zhang2016lock3dface}
Jinjin Zhang, Di Huang, Yunhong Wang, and Jia Sun.
\newblock Lock3dface: A large-scale database of low-cost kinect 3d faces.
\newblock In {\em 2016 International Conference on Biometrics (ICB)}, pages 1--8. IEEE, 2016.

\bibitem{zhao2022LMFNet}
Panzi Zhao, Yue Ming, Xuyang Meng, and Hui Yu.
\newblock Lmfnet: A lightweight multiscale fusion network with hierarchical structure for low-quality 3-d face recognition.
\newblock {\em IEEE Transactions on Human-Machine Systems}, 2022.

\bibitem{zhu2023exploiting}
Yizhe Zhu, Jialin Gao, Tianshu Wu, Qiong Liu, and Xi Zhou.
\newblock Exploiting enhanced and robust rgb-d face representation via progressive multi-modal learning.
\newblock {\em Pattern Recognition Letters}, 166:38--45, 2023.

\end{thebibliography}
}

\end{document}